\documentclass{article}
\usepackage[a4paper, total={6in, 8in}]{geometry}
\usepackage{hyperref}
\usepackage{graphicx}
\usepackage{tabularx,ragged2e}
\usepackage{algorithm,algpseudocode}
\usepackage{natbib}
\usepackage[symbol]{footmisc}

\usepackage{multirow}
\usepackage{amsfonts,mathtools}
\newcommand{\E}{\mathbb{E}}

\begin{document}
\begin{center}
	\LARGE PowerGym: A Reinforcement Learning Environment for Volt-Var Control in Power Distribution Systems 
\end{center}
\vspace{0.3cm}

\centerline{\large Ting-Han Fan\textsuperscript{\rm 1,3}, Xian Yeow Lee\textsuperscript{\rm 2,3}, and Yubo Wang \textsuperscript{\rm 3}} 
\vspace{0.3cm}
\centerline{\large \textsuperscript{\rm 1} Princeton University}
\centerline{\large \textsuperscript{\rm 2} Iowa State University}
\centerline{\large \textsuperscript{\rm 3} Siemens Technology}
\vspace{0.3cm}
\centerline{\large \texttt{tinghanf@princeton.edu, xylee@iastate.edu, yubo.wang@siemens.com}}
\vspace{0.3cm}

\begin{abstract}
We introduce PowerGym, an open-source reinforcement learning environment for Volt-Var control in power distribution systems. Following OpenAI Gym APIs, PowerGym targets minimizing power loss and voltage violations under physical networked constraints. PowerGym provides four distribution systems (13Bus, 34Bus, 123Bus, and 8500Node) based on IEEE benchmark systems and design variants for various control difficulties. To foster generalization, PowerGym offers a detailed customization guide for users working with their distribution systems. As a demonstration, we examine state-of-the-art reinforcement learning algorithms in PowerGym and validate the environment by studying controller behaviors. The repository is available at \url{https://github.com/siemens/powergym}.
\end{abstract}

\section{Introduction}
Volt-Var control refers to the control of voltage (Volt) and reactive power (Var) in power distribution systems to achieve healthy operation of the systems. By optimally dispatching voltage regulators, switchable capacitors, and controllable batteries, Volt-Var control helps to flatten voltage profiles and reduce power losses across the power distribution systems.  It is hence rated as the most desired function for power distribution systems ~\citep{borozan2001integrated}.

The center of the Volt-Var control is an optimization for voltage profiles and power losses governed by networked constraints. Represent a power distribution system as a tree graph ($\mathcal{N}, \xi$), where $\mathcal{N}$ is the set of nodes or \textit{buses} and $\xi$ is the set of edges or lines and transformers. 
Denote node $i$ as $j$'s parent. The physical networked constraints are given by \citep{Farivar2013voltage}:
\begin{equation}
\begin{split}
    p_j&=p_{ij}-R_{ij}\ell_{ij}-\sum_{(j,k)\in \xi}p_{jk}\\
    q_j&=q_{ij}-X_{ij}\ell_{ij}-\sum_{(j,k)\in \xi}q_{jk}\\
    v_j^2&=v_i^2-2(R_{ij}p_{ij}+X_{ij}q_{ij})+(R^2_{ij}+X^2_{ij})\ell_{ij}\\
    \ell_{ij}&=(p^2_{ij}+q^2_{ij})/ v_i^2,
\label{eq:pf-constraints}
\end{split}
\end{equation}
where $p,q$ are active and reactive power consumed at nodes or edges, $v,\ell$ denote bus voltage magnitude and squared current magnitude, $R,X$ are resistance and reactance. Capital letters stand for given parameters otherwise variables. The constraints in Eq.~\eqref{eq:pf-constraints} have quadratic equalities, making any optimization upon it nonconvex. Researchers have either tightly relaxed the constraints with strict nodal injection assumptions ~\citep{gan2014exact} or used linearization that assumes the distribution systems operates at a fixed operating point ~\citep{yang2016optimal}. Both methods require tremendous efforts in trimming and conversion of a model that is readily available in commercial circuit simulation software to specific optimization formulation. Together with many integer decision variables in controllable devices not shown above, the Volt-Var control problem becomes extremely hard to scale to a system with thousands of buses, a typical size for distribution systems. 

With recent breakthroughs in deep reinforcement learning (RL), power system researchers have tried using RL for power system operation. One such example is learning to operate a transmission systems operation in L2RPN competition~\citep{marot2021learning}. Though transmission systems are fundamentally different from distribution systems in both network topology (looped vs. radial) and typical problem types (dynamic vs. quasi-static), RL has shown promising results ~\citep{yoon2020winning} in operating transmission systems. While there exist many papers on RL in distribution systems, researchers have used their own environments. One of the many reasons behind it is due to the regulatory and conservative nature of the power engineering industry: being safety-critical, the real-life distribution system topologies and control settings are proprietary. To encourage power systems researchers to make fair comparisons on the developed RL algorithms without having the concern of proprietary information leakage, we have developed PowerGym, a Gym-like environment~\citep{brockman2016openai} for optimizing Volt-Var control using IEEE benchmark test systems~\citep{testfeeder,dugan2010ieee}. It further serves as a base for power systems engineers to implement RL algorithms on their proprietary systems with minor customization. 

PowerGym supports Gym-like usages such as reset, step, random action sampling, and visualization; hence it is readily applicable to run on most developed RL algorithms. On top of the Gym design, PowerGym provides a wide range of environment variations of the IEEE benchmark systems. These variations affect the environment's physical constraints and ultimately the control difficulties, allowing users to choose an environment either easier to control however more abstracted or harder to control yet more realistic. PowerGym is safe for parallel execution up to some file constraints (discuss in environment design section), so the user can run parallel algorithms such as A3C \citep{mnih2016a3c}.

Our contributions are as follows. First, we design PowerGym to help power system researchers benchmark their controls and RL algorithms. To be best of the authors’ knowledge, this is the first publicly accessible environment with a focus on Volt-Var control in power distribution systems. Second, we consider environment usability and extendibility in PowerGym. We provide variations of the environments for different control difficulties and a detailed customization guide. Finally, we showcase the applicability of PowerGym on two popular RL algorithms, PPO \citep{schulman2017ppo} and SAC \citep{haarnoja2018sac} for validation purposes. We also explain how the controllers work through a case study.

\section{Related Work}
The application of RL to control and manage various aspects of power systems is a well-studied topic in literature~\citep{DeepRLPowerApplicationsOverview}. There has been renewed interest in this topic due to algorithmic advancements, allowing RL to go beyond tabular settings and scale to large state and action spaces using neural networks as expressive function approximators. Examples of such work include home/building energy control~\citep{sun2020continuous,pigott2021gridlearn}, power systems stability control~\citep{1266597}, microgrid control~\citep{henri2020pymgrid}, and load frequency control for renewable energy~\citep{8534442}. In the context of Volt-Var control, various studies leverage RL to optimize various aspects of the problem, such as emphasizing the constraint satisfaction~\citep{8909741, 8944292} or the sample efficiency and scalability~\citep{9143169}. Nevertheless, most of these results are based on non-standardized implementations of various systems with the environment tuned to the specifics of the problem. This has led to the difficulty of evaluating and comparing results and remains a crucial challenge in these areas, as highlighted by~\citet{DeepRLPowerChallenges}.

In other domains of deep RL, researchers recognized the importance of having high-quality benchmark environments to facilitate the research into RL application. This has led to the development of environments such as OpenAI Gym for training RL agents to play a variety of games and for robotic control~\citep{brockman2016openai}, Safety Gym for safe RL exploration~\citep{Ray2019} and ns3-gym for training RL in networks research problems~\citep{gawlowicz2018ns3}. More closely related benchmarks include the Grid2Op \citep{grid2op} and Gym-ANM \citep{henry2021anm}. Grid2Op is the platform used for the Learning To Run Power Network (L2PRN) challenge, controlling unpredictable power generation from renewable energy sources through power systems' topology change~\citep{MAROT2020106635}. Gym-ANM is an RL environment that models Active Network Management (ANM) for renewable energy. Nevertheless, a standardized benchmark for Volt-Var control environments with the flexibility of instantiating systems of various sizes and difficulty is still lacking. To this end, we hope that PowerGym serves to fill the gap in the community as a unified benchmark environment for RL research in Volt-Var control.

\section{Reinforcement Learning Preliminaries} 
A reinforcement learning (RL) environment is often modeled by the Markov Decision Process (MDP) with two common MDPs: infinite-horizon discounted MDP $M_d=\langle \mathcal{S}, \mathcal{A}, T, r, \gamma \rangle$ and finite-horizon episodic MDP $M_e=\langle \mathcal{S}, \mathcal{A}, \{T_i\}, \{r_i\}, H \rangle$. $\mathcal{S}$, $\mathcal{A}$ are the state and action spaces. $T$/$\{T_i\}$, $r$/$\{r_i\}$ are the stationary/non-stationary state transition function and reward function. $\gamma\in(0,1)$, $H\in\mathbb{N}$ are the discount factor and the horizon. In this paper, we assume a stationary state transition in $M_e$: $T_i=T$ for $i\in[0...H-1]$. The goal of RL is to find a policy to maximize the cumulative rewards:
\begin{equation}
\begin{split}
    &R_{M_d}(\pi)=\E\Big[\sum_{i=0}^\infty \gamma^i r(s_i,a_i,s_{i+1})\Big|a_i\sim\pi(\cdot|s_i) \Big]\\
    &R_{M_e}(\{\pi_i\})=\E\Big[\sum_{i=0}^{H-1} r_i(s_i,a_i,s_{i+1})\Big|a_i\sim\pi_i(\cdot|s_i) \Big]
\end{split}
\label{eq:cum-reward}
\end{equation}
It is well-known that the optimal policy of $M_d$ is stationary while that of $M_e$ is non-stationary \citep{Agarwal2019ReinforcementLT}[Chapter 1]; hence in Eq.~\eqref{eq:cum-reward}, the policy is denoted as $\pi$ in $M_d$ and as $\{\pi_i\}$ in $M_e$. To reduce the model complexity, most RL experiments are formulated into $M_d$ if the stationarity holds. However, $M_e$ is inevitable when the reward is non-stationary. Depending on the application scenarios, we implement both stationary and non-stationary rewards, which will be discussed in the next section. 

\section{A Volt-Var Control Environment} 
\subsection{Power Distribution Systems and Objectives}

Power distribution systems are networks for delivering electric power from the power transmission system to end consumers. Due to the distribution loss, voltage drops along the power delivery line, possibly causing voltage violations and power losses. Thus, Volt-Var optimization is required. In power distribution systems, the Volt-Var optimization problem is to control devices (e.g., regulators, capacitors, and batteries. represented as $x$.) under constraints. $x$ affects voltage, resistance, reactance, and power in the physical networked constraints, so Eq.~\eqref{eq:pf-constraints} is a constraint of Eq.~\eqref{eq:voltvar}. 
\begin{equation}
\begin{matrix*}[l]
\min_x &  f_{\text{volt}}(x)+f_{\text{ctrl}}(x)+f_{\text{power}}(x)\\
\text{s.t.}& \text{Eq}.~\eqref{eq:pf-constraints}~\text{and~device~constraints.}
\end{matrix*}
\label{eq:voltvar}
\end{equation}
The Volt-Var optimization's objective is a combination of three losses: $f_{\text{volt}}$ for voltage violation, $f_{\text{ctrl}}$ for control error, and $f_{\text{power}}$ for power loss. The device constraints ensures the devices operates within its physical limits. While Eq.~\eqref{eq:voltvar} only accounts for a single time step, in practice we solve it at every time step. Solving a sequence of Volt-Var optimization, Eq.~\eqref{eq:voltvar}, becomes a Volt-Var control problem. In short, we call a problem \emph{Volt-Var optimization} if solving a single Eq.~\eqref{eq:voltvar} and \emph{Volt-Var control} if solving a sequence of Eq.~\eqref{eq:voltvar} connected by device operation constraints over time. 

We use a Python version of OpenDSS to solve for the physical networked constraints of Eq.~\eqref{eq:voltvar}. OpenDSS is an open-source power flow solver developed by EPRI. It takes $p,q$ from Eq.~\eqref{eq:pf-constraints} as known and solves the nonlinear equations of voltages and currents using fixed-point iterations.

\begin{figure}[!ht]
    \centering
    \includegraphics[width=0.8\textwidth]{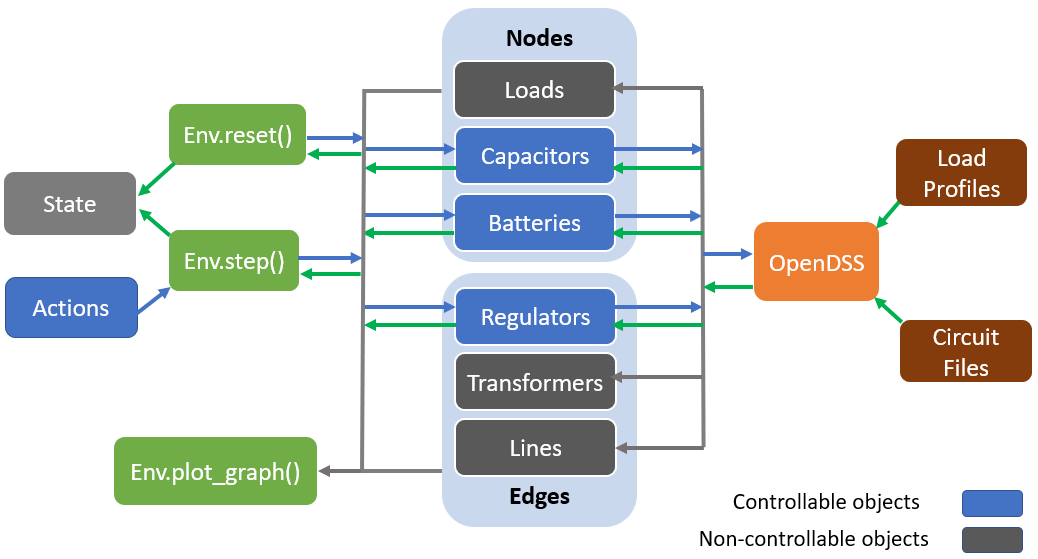}
    \caption{A depiction of PowerGym. OpenDSS reads the circuit files and load profiles (introduce in environment registration) to initialize the simulation. The environment functions interact with OpenDSS to get the state transitions.}
    \label{fig:data-flow}
\end{figure}

Shifting the focus to elements in power distribution systems, we define each element to be multi-phase following the fact that power is usually delivered in multi-phases. As shown in Figure~\ref{fig:data-flow}, a (multi-phase) node, or a bus, can be a pure connection point or include node objects like loads, capacitors, or batteries. A (multi-phase) edge is formed by a line, a transformer, or a regulator. Loads model the power consumption from the consumers. Capacitors provide reactive power and batteries are energy (active power) storage. Lines imitate the connection from one (multi-phase) node to another subject to Ohm's law. Transformers and regulators are for voltage adjustment from one node to another.

\subsection{Volt-Var Control as An RL Problem}

In this subsection, we describe the Volt-Var control problem in the language of RL. We consider a finite-horizon MDP with $H=24$ steps as the horizon because we focus on a daily control with the control frequency being one action per hour. Still, the horizon is a variable and hence changeable. We will discuss this in the sections of environment registration and experiments. The following paragraphs give the details about the observation space, the action space, the state transition, and the reward function in PowerGym.

\subsubsection{Observation and Action Spaces}
The observation and action spaces, as summarized in Table~\ref{tbl:obs_space} and \ref{tbl:act_space}, are products of discrete and continuous variables. The discrete variables are from the physical constraints of the controllers; for example, a capacitor either turns on or off, a regulator operates on a finite number of modes (tap number), and a discrete battery only has a finite number of discharge powers. The continuous variables are normalized into some bounded ranges; for example, the (per-unit) voltage is represented into the unit of the base voltage on a bus, and hence usually bounded in [0.8,~1.2]. The battery's state-of-charge (charge / max charge), or the soc, is in [0.0,~1.0]. The continuous battery's normalized discharge power (discharge power / max discharge power) is in [-1.0,~1.0], where the negative means charging and the positive means discharging.

Depending on the device constraints, a battery control can be either discrete or continuous. This affects the action space (Table~\ref{tbl:act_space}) since the action representations are different. Still, because we can post-process the observation after receiving it, in the observation space (Table~\ref{tbl:obs_space}), we unify the representation of discrete and continuous batteries by mapping the discrete battery's discharge power to the normalized form. 

Whether to discretize the battery makes the action space either multi-discrete or a product of multi-discrete and continuous spaces. Either way, the problem is hard for a tabular policy or a policy that encodes the actions as one-hot vectors (e.g., DDQN \citep{van2016ddqn}) because the size of the discrete part of action space scales exponentially in the number of controllers. Also, the possibility of mixing discrete and continuous actions makes it harder to design the policy.
\begin{table}[!ht]
    \centering
    \begin{tabular}{l|ll}
        \hline
        \hline
        \textbf{Variable} & \textbf{Type} & \textbf{Range} \\ \hline
        Bus voltage & cont. & $[0.8,1.2]$ \\
        Capacitor status & disc. & $\{0,1\}$ \\
        Regulator tap number & disc. & $\{0,...,N_{\text{reg\_act}}-1\}$ \\
        State-of-charge (soc) & cont. & $[0,1]$ \\
        Discharge power & cont. & $[-1,1]$ \\ \hline
        \hline
    \end{tabular}
    \caption{Observation Space. Here, both discrete and continuous batteries give the normalized discharge power. $N_{\text{reg\_act}}$ is the number of taps of a regulator (default value: 33).}
    \label{tbl:obs_space}
\end{table}
\begin{table}[!ht]
    \centering
    \begin{tabular}{l|ll}
        \hline
        \hline
        \textbf{Variable} & \textbf{Type} & \textbf{Range} \\ \hline
        Capacitor status & disc. & $\{0,1\}$ \\
        Regulator tap number & disc. & $\{0,...,N_{\text{reg\_act}}-1\}$ \\
        Discharge power (disc.) & disc. & $\{0,...,N_{\text{bat\_act}}-1\}$ \\
        Discharge power (cont.) & cont. & $[-1,1]$ \\ \hline
        \hline
    \end{tabular}
    \caption{Action Space. The discrete battery uses the discretized discharge power while the continuous battery uses the normalized one. $N_{\text{bat\_act}}$ is the number of a discrete battery's discharge power (default value: 33).}
    \label{tbl:act_space}
\end{table}

\subsubsection{State Transition}
We now describe the state transition function $s' = T(s,a)$ in PowerGym. With the descriptions in Table~\ref{tbl:obs_space} and \ref{tbl:act_space},~$s'$ is represented as
\begin{equation}
s'=[\text{Vols}(s,a),~\text{cap}(a),~\text{reg}(a),~\text{soc}(s,a),~\text{dis}(s,a)].
\end{equation}
$\text{Vols}(s,a)$ is the next set of voltages and depends on action $a$ and the stochasticity of loads, which we model using the load profiles (will discuss in environment design). $\text{cap}(a)$ and $\text{reg}(a)$ are the next statuses of capacitors and regulators. $\text{soc}(s,a)$ and $\text{dis}(s,a)$ are the next soc's and discharge powers of batteries. Both of them depend on the current state $s$ because a battery's soc cannot go beyond full charge ($\text{soc}=1$) or depleted ($\text{soc}=0$). To enforce this, we project the \emph{attempted} discharge power in an action $a$ to the allowed range based on $s$, making $\text{dis}(s,a)$ a function of $(s,a)$.

\subsubsection{Reward Function}
We implement the objective of a Volt-Var problem, Eq.~\eqref{eq:voltvar}, into a reward function as follows:

\begin{equation}
	r(s,s',i) = -f_{\text{volt}}(s') - f_{\text{ctrl}}(s,s',i) -f_{\text{power}}(s')
\label{eq:reward_func}
\end{equation}
\begin{equation}
    f_{\text{power}}(s')= \text{w}_{\text{power}}\frac{\text{PowerLoss}(s')}{\text{TotalPower}(s')}
\label{eq:power_loss}
\end{equation}
$s$ is a concatenation of all observations in the current step, $s'$ is that in the next step, and $i\in[0,...,H-1]$ is the episode step. The dependency on step $i$ implies the reward could be non-stationary. The power loss, Eq.~\eqref{eq:power_loss}, is a ratio of the overall power loss to the total power. The voltage violation and control error are expressed in Eq.~\eqref{eq:vol-violation} and \eqref{eq:ctrl-err}. 

Eq.~\eqref{eq:reward_func} is expressed as $r(s,s',i)$, not $r(s,a,i)$, because the action $a$ is a part of the next state $s'$. Mathematically, $r(s,a,i)$ and $r(s,s',i)$ are equivalent because $s'=T(s,a)$ is a function of $(s,a)$ under the state transition function $T$.

The voltage violation, Eq.~\eqref{eq:vol-violation}, is a sum of worst-case voltage violations among all phases across all the nodes in the system. The upper/lower violation thresholds ($\overline{V}$/$\underline{V}$) are set as $\pm5\%$ of the per-unit voltage $V$ as a result of the US voltage regulation standard \citep{VolStandard}.
\begin{equation}
\begin{split}
    f_{\text{volt}}(s')= &\sum_{n\in \mathcal{N}} (\max_{p\in \text{Phases}(n)}V_{n,p}(s')-\overline{V})_+\\
	&+(\underline{V}-\min_{p\in \text{Phases}(n)}V_{n,p}(s'))_+,
\end{split}
\label{eq:vol-violation}
\end{equation}
where $(\cdot)_+ $ is a shorthand for $ \max(\cdot,0)$. Thereby, the upper violation $(\max_p V_{n,p} - \overline{V})_+$ is positive when $\max_p V_{n,p}>\overline{V}$ and zero otherwise.

The control error, Eq.~\eqref{eq:ctrl-err}, is a sum of capacitors' and regulators' switching penalties ($1^\text{st}$ \& $2^\text{nd}$ rows) and batteries' discharge penalty and soc penalty ($3^\text{rd}$ row). These penalties discourage the policy from making frequent changes and slow the devices from wear out. Note the discharge error $\frac{\text{P}_b(s')_+}{\overline{\text{P}_b}}$, with $\overline{\text{P}_b}$ being the max power, has a $(\cdot)_+$ function as the battery degradation is primarily caused by the battery discharging power $\text{P}_b>0$. Besides, the soc penalty has an indicator of the last time step $\mathbb{I}_{i=H}$ to encourage a battery $b$ to return to its initial state-of-charge $\text{soc0}_b$. Hence, the reward is stationary if $\text{w}_{\text{soc}}=0$ and non-stationary otherwise.
\begin{equation}
\begin{split}
    &f_{\text{ctrl}}(s,s',i)=\sum_{c\in \text{caps}}\text{w}_{\text{cap}}|\text{Status}_c(s)-\text{Status}_c(s')|\\
    &\quad\quad\quad\quad\quad\quad+\sum_{r\in \text{regs}}\text{w}_{\text{reg}}|\text{TapNum}_r(s)-\text{TapNum}_r(s')|\\
	&\quad\quad\quad\quad\quad\quad+\sum_{b\in \text{bats}}\text{w}_{\text{dis}}\frac{\text{P}_b(s')_+}{\overline{\text{P}_b}}+\text{w}_{\text{soc}}\mathbb{I}_{i=H}|\text{soc}_b(s')-\text{soc0}_b|,  
\end{split}
\label{eq:ctrl-err}
\end{equation}
where $c,~r,~b$ represent a capacitor, a regulator, and a battery. $\text{Status}_c$, $\text{TapNum}_r$, $\text{P}_b$,  $\text{soc}_b$ are status of $c$, tap number of $r$, discharge power of $b$, and soc of $b$.

\section{Design of PowerGym}
\subsection{Environment Instantiation}

Similar to the OpenAI Gym, PowerGym provides \texttt{make\char`_env()} to instantiate an environment:
\begin{quote}
\begin{verbatim}
make_env(env_name, worker_idx=None)
\end{verbatim}
\end{quote}
\texttt{env\char`_name} is the name of the registered environment. \texttt{worker\char`_idx} is used (if not \texttt{None}) for parallel execution, which we detail in the subsection of load profiles.

\texttt{make\char`_env()} reads the following information. First, PowerGym reads circuit files into the environment class, followed by leveraging OpenDSS to compile the file, as shown in Figure~\ref{fig:data-flow}. Secondly, to define the hyper-parameters that affects the RL training under the same system, PowerGym needs information such as the horizon, the number of actions of a regulator/battery and weights of the power loss, capacitor's switch loss, regulator's switch loss, battery's discharge loss, battery's state-of-charge (soc) loss. The next subsection introduces the customization of such information.

\subsection{Environment Registration and Customization}
Users can customize their environment by registering a new environment name associated with the required information. This is done by appending the information to the dictionaries in the PowerGym register. Below is an example of the dictionary. \texttt{dss\char`_file} is the main circuit file that OpenDSS compiles. Users can edit \texttt{dss\char`_file} to change the circuit objects and structure. Users can also change the hyper-parameters. \texttt{max\char`_episode\char`_steps} is the horizon. It is 24 by default as we focus on the daily control. \texttt{act\char`_num} is the shorthand of the number of actions, so the battery becomes continuous if \texttt{bat\char`_act\char`_num} is infinity and discrete if finite. The other parameters are the weights in the reward function shown in Eq.~\eqref{eq:reward_func}.
{\small
\begin{quote}
\begin{verbatim}
'13Bus': {
  'system_name': '13Bus',
  'dss_file': 'IEEE13Nodeckt_daily.dss',
  'max_episode_steps': 24,
  'reg_act_num': 33,
  'bat_act_num': 33,
  'power_w': 10.0,
  'cap_w': 1.0/33,
  'reg_w': 1.0/33,
  'soc_w': 0.0/33,
  'dis_w': 6.0/33 }
\end{verbatim}
\end{quote}
}
Besides the information shown above, PowerGym also depends on the load profiles (see Figure~\ref{fig:data-flow} and the subsequent subsection) and the other circuit files. These files are customizable and can be found in the folder of \texttt{systems/system\char`_name} of the repository. For example, the above shows users can find customizable files in the folder \texttt{systems/13Bus}.

\subsection{Default Registered Environments}

\begin{table}[!ht]
    \centering
    \begin{tabular}{l|l}
        \hline
        \hline
        System & Environment Names \\ \hline
        \multirow{2}{*}{13Bus} & \texttt{13Bus},~\texttt{13Bus\char`_cbat},~\texttt{13Bus\char`_soc}\\ & \texttt{13Bus\char`_cbat\char`_soc} 
        \\ \hline
        \multirow{2}{*}{34Bus} & \texttt{34Bus},~\texttt{34Bus\char`_cbat},~\texttt{34Bus\char`_soc}\\ & \texttt{34Bus\char`_cbat\char`_soc} 
        \\ \hline
        \multirow{2}{*}{123Bus} & \texttt{123Bus},~\texttt{123Bus\char`_cbat},~\texttt{123Bus\char`_soc}\\ & \texttt{123Bus\char`_cbat\char`_soc} 
        \\ \hline
        \multirow{2}{*}{8500Node} & \texttt{8500Node},~\texttt{8500Node\char`_cbat}\\ & \texttt{8500Node\char`_soc},~\texttt{8500Node\char`_cbat\char`_soc} 
        \\ \hline
        \hline
    \end{tabular}
    \caption{Environment is named in \texttt{\{system name\}\char`_\{use continuous battery\}\char`_\{use soc penalty\}}. }
    \label{tbl:envs}
\end{table}
In Table~\ref{tbl:envs}, each system (summarized in Table~\ref{tbl:sys}) in PowerGym has four default environments: vanilla, continuous battery, soc, continuous battery \& soc. The difference lies only in the battery's settings; hence capacitors and regulators are the same across these four environments. 

The presence of \texttt{cbat} affects the battery's number of discharge power: without \texttt{cbat}, the number is finite (33 by default), and the battery's model is discrete; with \texttt{cbat}, the number is infinite, and the battery's model is continuous. On the other hand, \texttt{soc} tells the state-of-charge penalty on the battery at the end of the horizon: without \texttt{soc}, the soc penalty is zero, and the reward is stationary; with \texttt{soc}, the soc penalty is positive, and the reward is non-stationary. Besides the four default environments, one can call a scaled environment by appending a scale to an environment name; e.g., \texttt{13Bus\char`_s1.5} scales the loads by 1.5. We will revisit the load scaling in the subsection of load profiles.

\begin{table}[!ht]
    \centering
    \begin{tabular}{l|lll}
        \hline
        \hline
        System  & \# Caps & \# Regs & \# Bats\\ \hline
        13Bus  & 2 & 3 & 1\\
        34Bus  & 2 & 6 & 2\\
        123Bus  & 4 & 7 & 4\\
        8500Node  & 10 & 12 & 10\\ \hline
        \hline
    \end{tabular}
    \caption{System specifications. The system layouts can be found in the Appendix.}
    \label{tbl:sys}
\end{table}

\subsection{Gym-like Usage}
PowerGym supports Gym-like usages such as reset, step, random action sampling, and visualization. The design is compatible with most RL algorithms. Below is a brief overview of these functions.

\begin{quote}
\begin{verbatim}
obs = Env.reset(load_profile_idx=0)
\end{verbatim}
\end{quote}
The reset function initializes the system and returns an initial observation. The dynamics of the load are controlled by the load profile index and will be discussed in the next subsection. The initial statuses of capacitors, regulators, and batteries are set as "on", full tap number, and (full charge, zero discharge power), respectively.

{\small
\begin{quote}
\begin{verbatim}
obs, reward, done, info = Env.step(action)
\end{verbatim}
\end{quote}
}
The step function takes an action as the input and returns the next state, reward, done signal, and the information dictionary. Since the current design does not define a terminal state that should be strictly avoided, the done signal is true only when the episode step reaches the horizon. The information dictionary includes several details about the reward such as capacitor error, regulator error, discharge error, soc error, and soc (all in average), which also facilitates the application of multi-objective RL \citep{liu2014multiobjective}.

\begin{quote}
\begin{verbatim}
action = Env.random_action()
\end{verbatim}
\end{quote}
Random actions can also be generated by \texttt{Env.action\char`_space.sample()} and the random seed is set by \texttt{Env.seed()}. The action is a ($N_{\text{cap}}+N_{\text{reg}}+N_{\text{bat}}$)-dimensional array for the control signal on the controllers, with $N$ being the number of a certain controller.

\begin{quote}
\begin{verbatim}
fig, pos = Env.plot_graph()
\end{verbatim}
\end{quote}
The plot graph function returns a Matplotlib figure and a dictionary of node positions for users to visualize the network status. It supports options such as \texttt{show\char`_voltages}, \texttt{show\char`_controllers}, and \texttt{show\char`_actions}.

\subsection{Other Usages and Constraints of Load Profiles}

To simulate load dynamics, OpenDSS supports time-series simulations following some predefined load curves. A group of predefined curves of all loads is called a load profile. As mentioned in the previous subsection, \texttt{Env.reset()} has an option of load profile selection. Hence, PowerGym models the stochasticity of state transition using the load profiles.

By enlarging the values in the load profile with a fixed scale, PowerGym creates environments with various load scales. Since power consumption scales linearly with the load scale, the environment tends to be hard under a large load scale. Referring to the subsection of default environments, a scaled environment is instantiated by appending a scale to an environment name. During the call, PowerGym generates the load profiles under the corresponding scale factor and another text file to store the current load scale. Note the load profile is regenerated only if the previous load scale (stored in the text file) is different from the current one.

Due to the file dependency on the load profile, parallel execution is possible under certain conditions. As mentioned earlier, the worker index of \texttt{make\char`_env()} is used for parallel execution. When it is \texttt{None}, PowerGym cannot execute two environments on the same system (e.g., cannot execute \texttt{13Bus}, \texttt{13Bus\char`_cbat} together) due to the conflict of load profile selection. This is solved when the worker index is an integer because each worker has a distinct profile selection file. However, even with the worker index, PowerGym cannot execute environments in parallel with names that differ only in the load scales (e.g., \texttt{13Bus\char`_s1.0}, \texttt{13Bus\char`_s2.0}) because it only allows one load scale at any given time.

\section{Experiments}
\subsection{Cumulative Rewards in Default Environments}

\begin{figure}[!ht]
    \centering
    \includegraphics[width=0.4\textwidth]{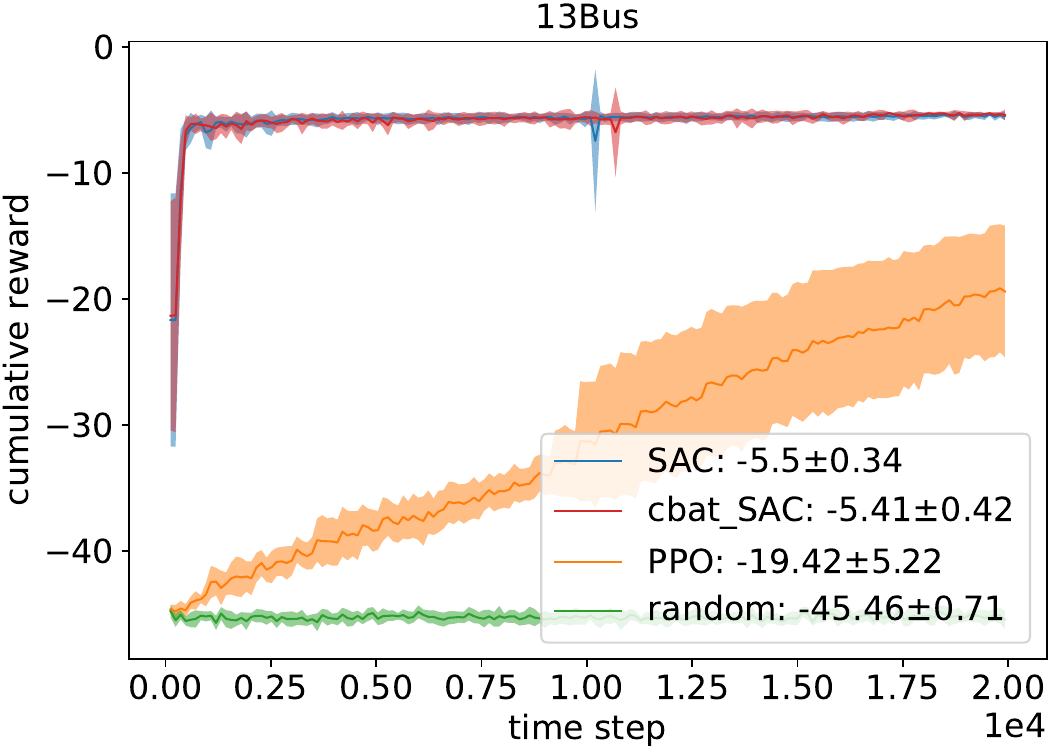}
    \includegraphics[width=0.4\textwidth]{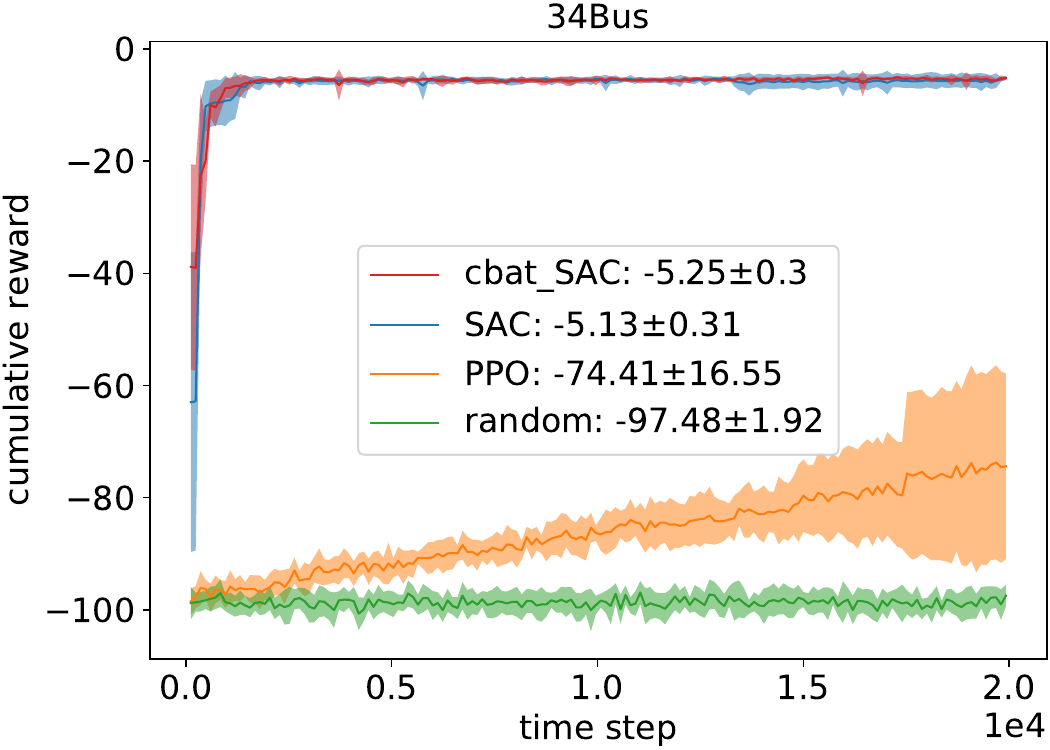}
    \includegraphics[width=0.4\textwidth]{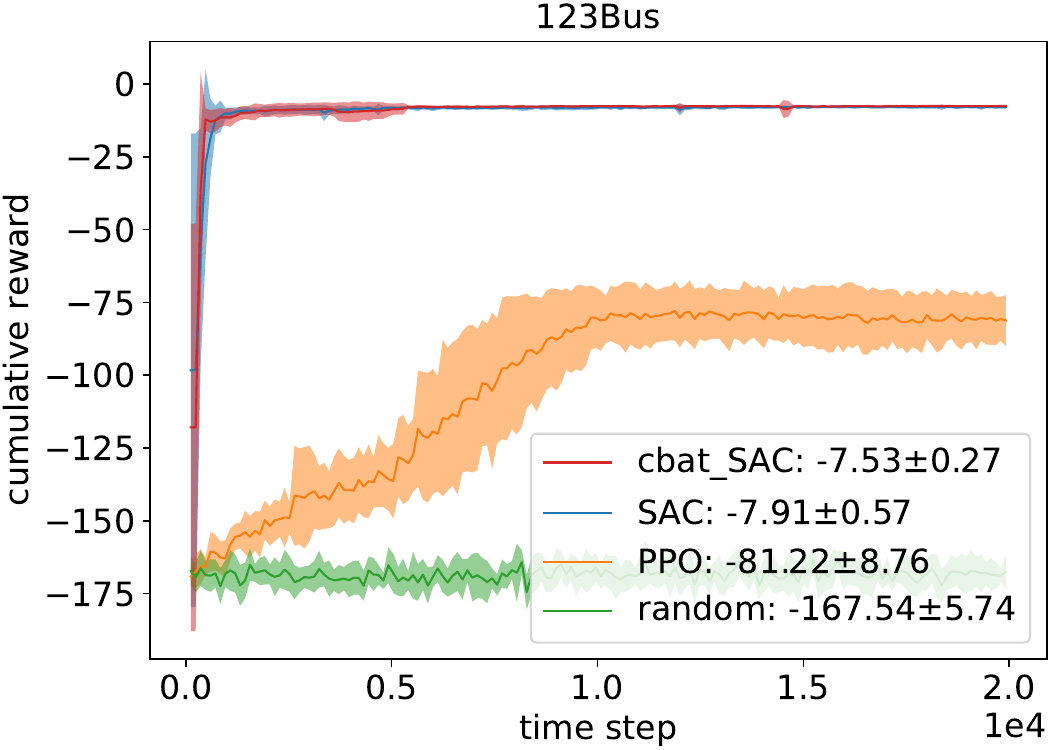}
    \includegraphics[width=0.4\textwidth]{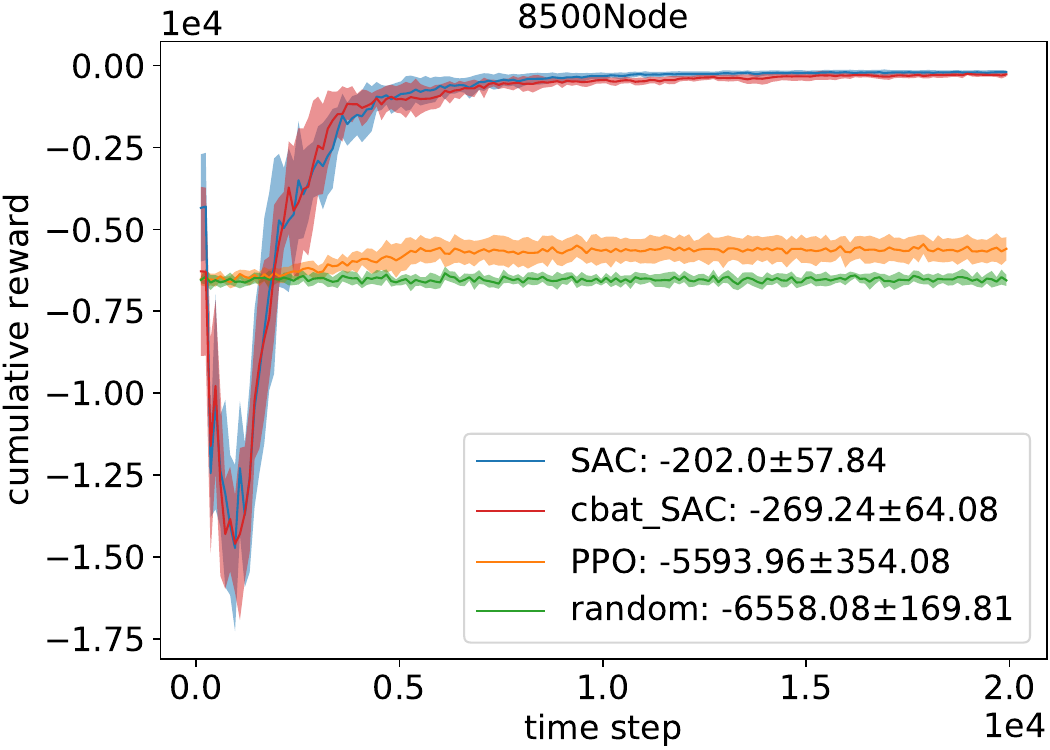}
    \caption{Cumulative rewards on test load profiles for different agents in 40k steps. The labels denote the average and standard deviation of the final rewards.}
    \label{fig:cum_reward}
\end{figure}

To show the applicability of PowerGym, we have trained two popular RL algorithms as benchmarks on our environments: Proximal Policy Optimization (PPO) \citep{schulman2017ppo} and Soft Actor-Critic (SAC) \citep{haarnoja2018sac}, with the implementations of PPO based on \citet{Fujita2021pfrl} and SAC on \citet{fan2021sac-integer}. Since PPO is on-policy while SAC is off-policy, these two algorithms give us a proxy of the expected performance of on-policy versus off-policy algorithms in the environments. For comparisons, both PPO and SAC have been trained on multi-discrete actions by default. In addition, SAC has been trained on environments with continuous batteries (cbat) to compare the environments with different battery settings. The experiments are run on a server with one AMD Ryzen Threadripper 3970X CPU and one Nvidia RTX 3090 GPU.

The experiments have been designed as follows: The load profiles are randomly partitioned into two halves, one for training and the other for testing. During training, the policy is tested on test load profiles every 5 episodes; or equivalently every 120 steps as the horizon is 24. Lastly, all experiments are performed across ten random seeds.

In Figure~\ref{fig:cum_reward}, the label "random" denotes an untrained policy that samples actions uniformly from the action space. As expected, SAC converges faster and outperforms PPO across all environments, which aligns with the SAC paper \citep{haarnoja2018sac} that has been demonstrated on the MuJoCo \citep{todorov2012mujoco}. Due to the experiment design (evaluation at every 120 steps), all curves start at step 120 instead of step 0. The first evaluation (step 120) reveals the algorithms' performance based on the first few updates: PPO is similar to random policy while SAC isn't. The fact that PPO is near the random policy validates the clipping nature of its policy gradient. Clipping makes PPO update slowly yet steadily and hence similar to random policy in the early steps. As for SAC, because its DDPG-style policy gradient \citep{silver2014ddpg} isn't clipped, SAC suffers more from the initial inaccuracy of Q-function and hence deviates from the random policy in the early steps. Finally, the performances of \texttt{SAC} and \texttt{cbat\char`_SAC} are very similar, implying discrete and continuous batteries share similar behaviors, and SAC successfully adapts to both.

To sum up, we have demonstrated the applicability of PPO and SAC in PowerGym and the sample efficiency of SAC. We have also shown that environments with discrete or continuous batteries have similar performances.

\subsection{Case Study: 123Bus}

\begin{figure}[!ht]
    \centering
    \includegraphics[width=0.8\textwidth]{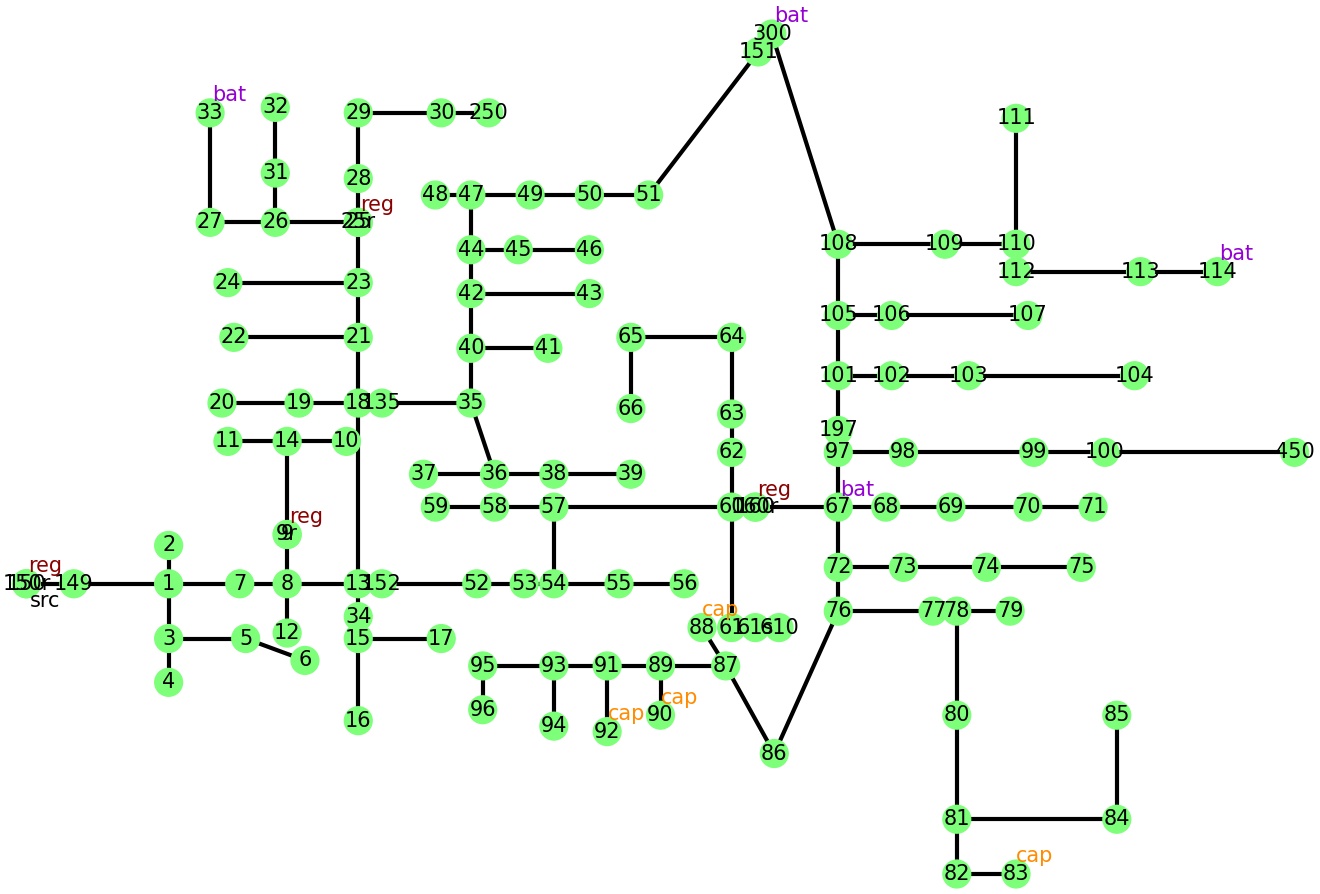}
    \caption{Layout of 123Bus system}
    \label{fig:123Bus_layout}
\end{figure}

\begin{figure}[!ht]
    \centering
    \includegraphics[width=0.4\textwidth]{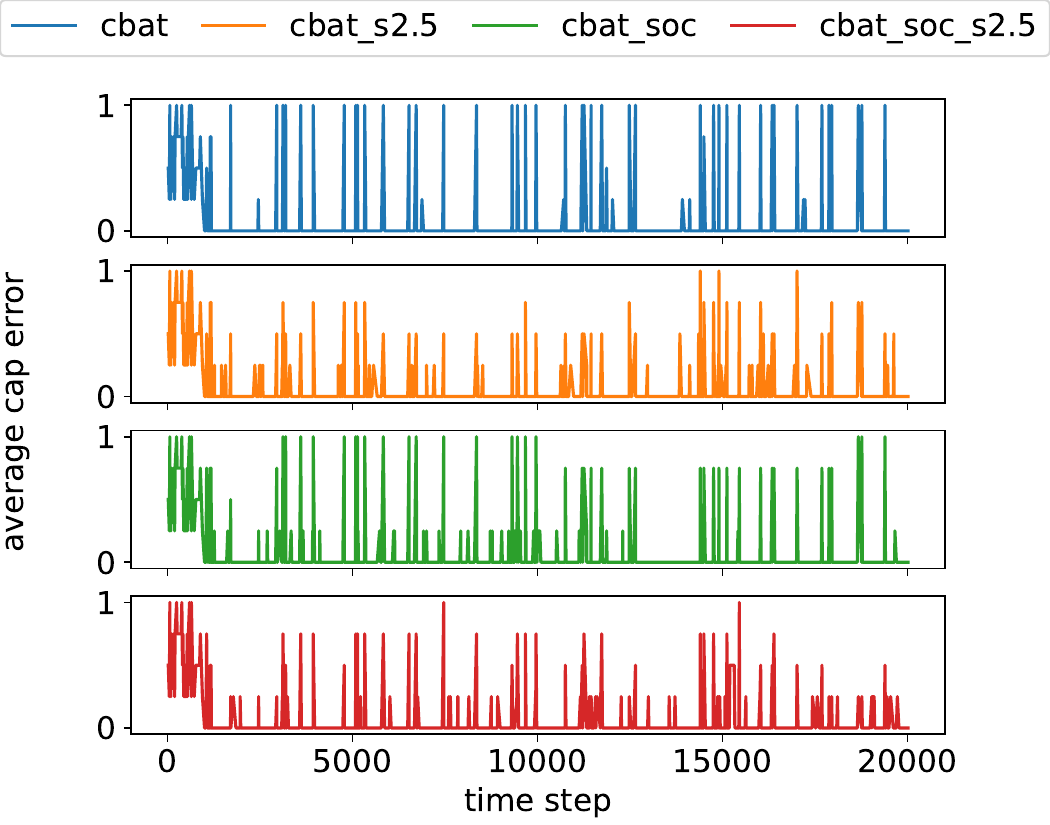}
    \includegraphics[width=0.4\textwidth]{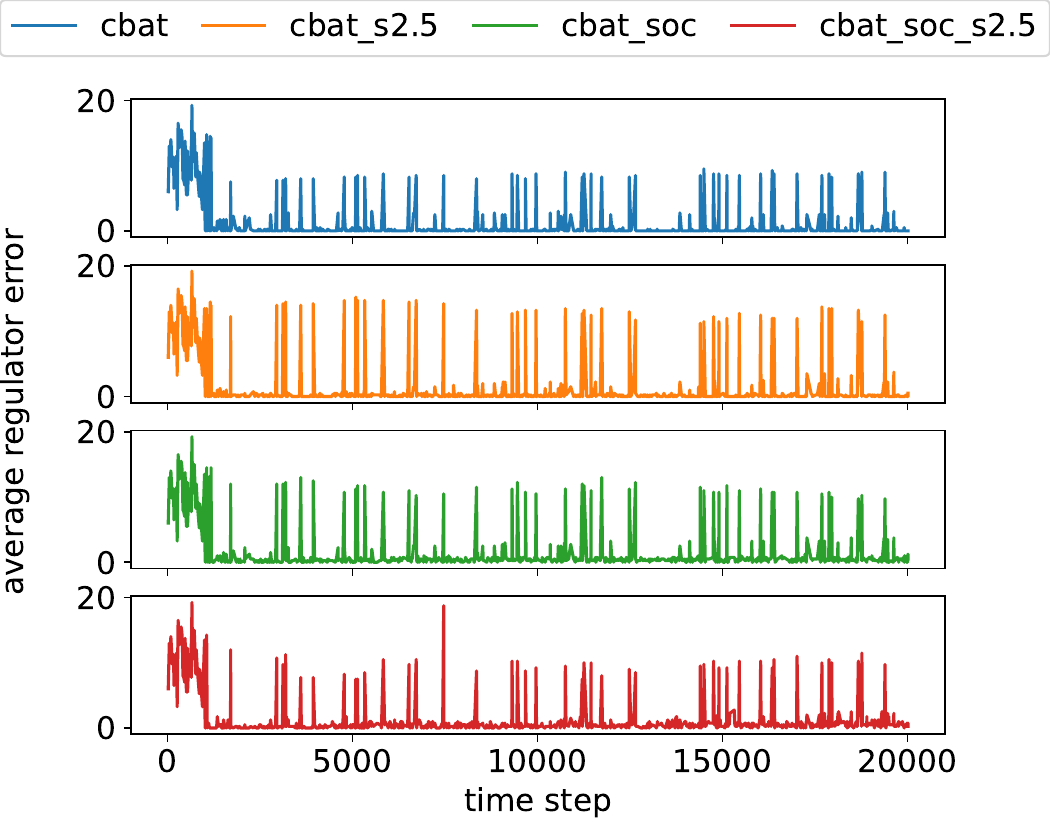}
    \includegraphics[width=0.4\textwidth]{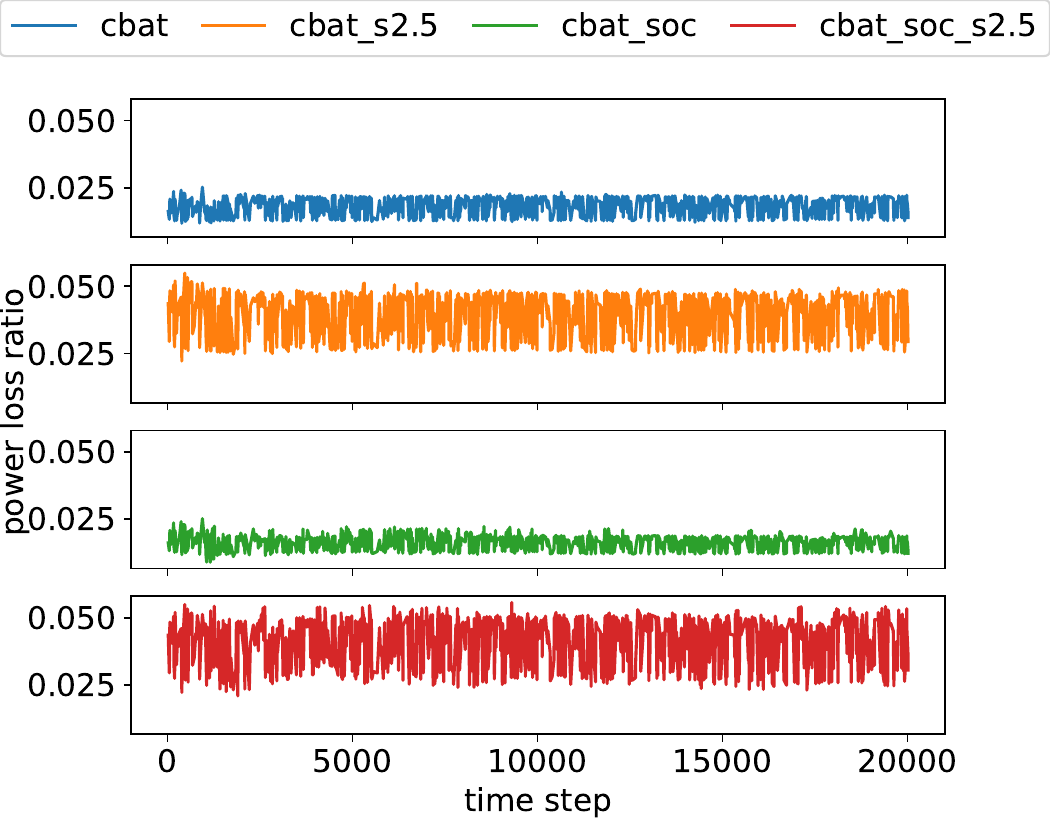}
    \includegraphics[width=0.4\textwidth]{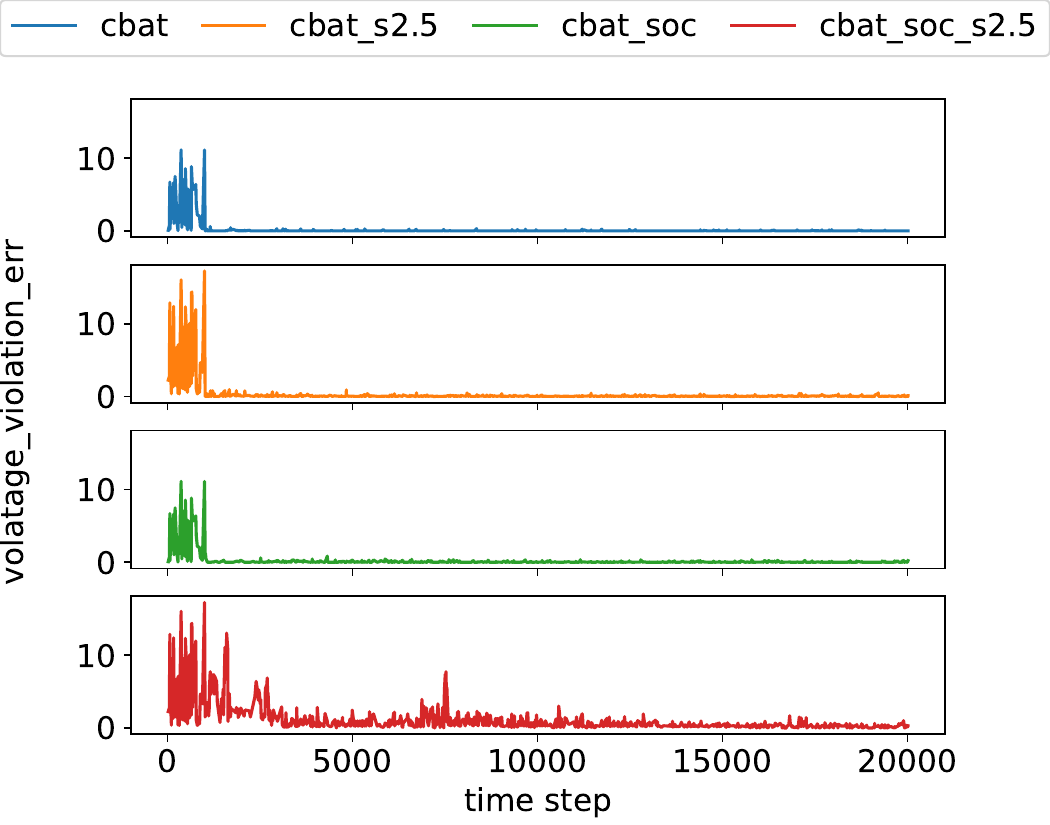}
    \includegraphics[width=0.4\textwidth]{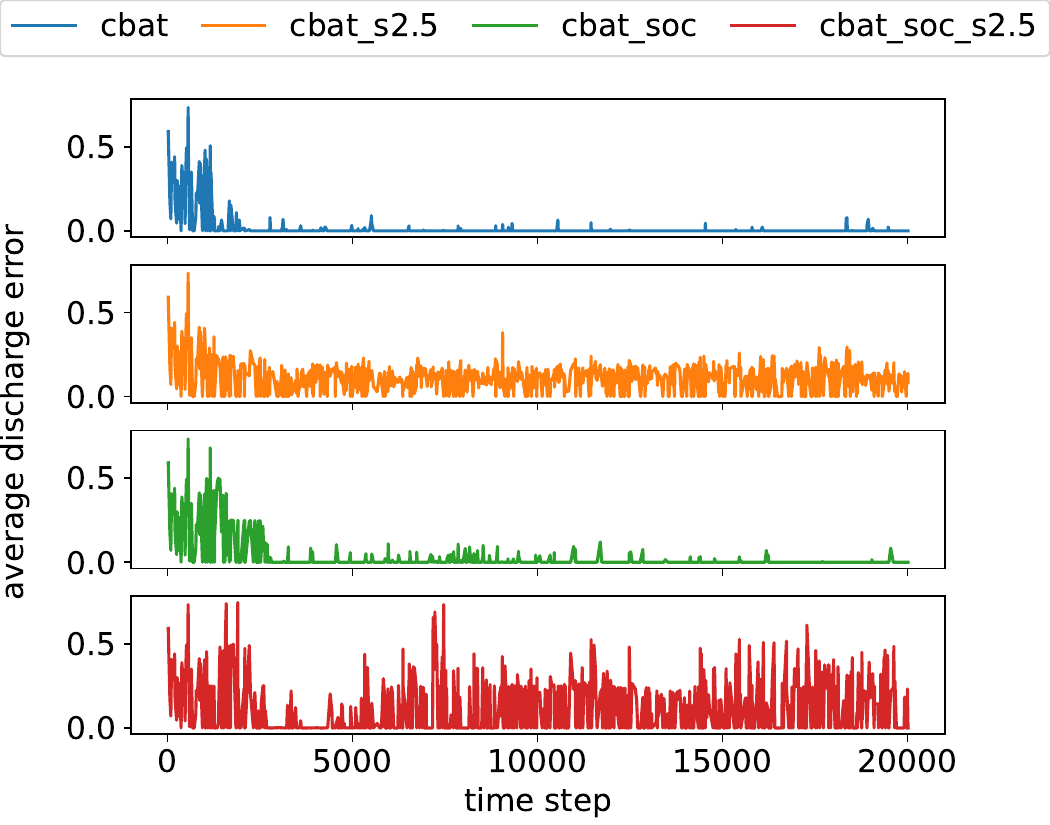}
    \includegraphics[width=0.4\textwidth]{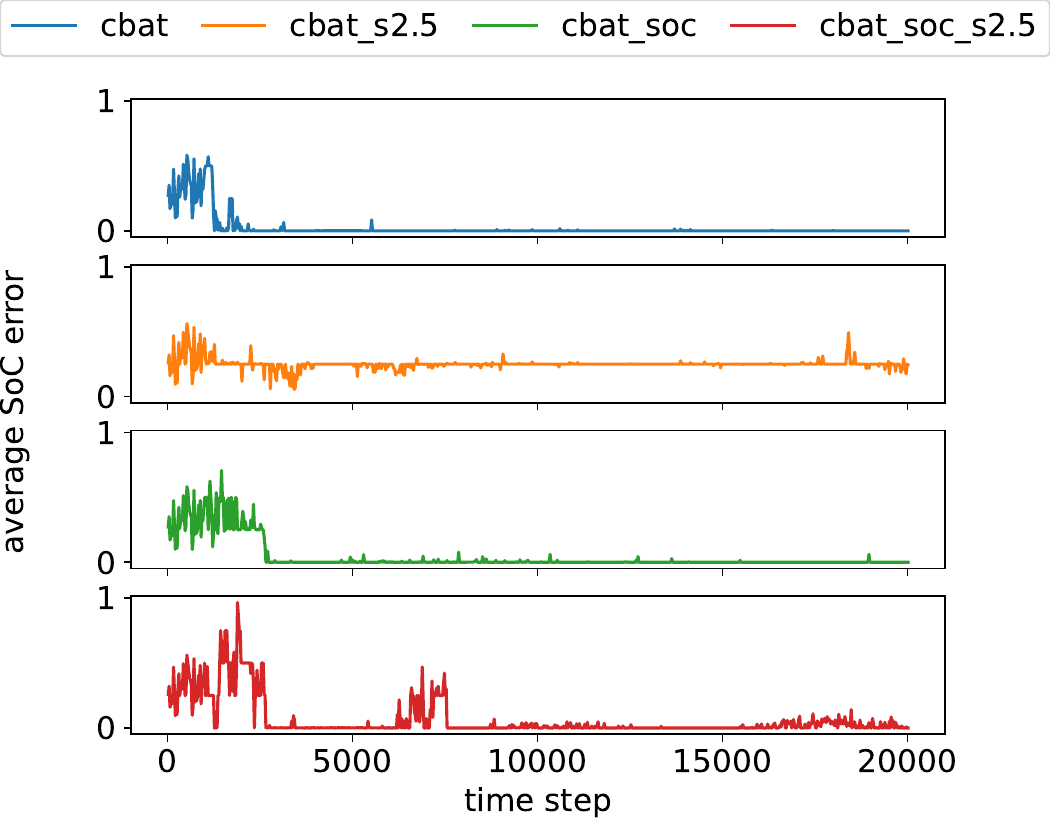}
    \caption{Immediate Errors on capacitors, regulators, power, voltage, discharge, and soc for \texttt{123Bus\char`_cbat}.}
    \label{fig:123bus_err}
\end{figure}
We take the 123Bus system (Figure~\ref{fig:123Bus_layout}) as an example to further analyze the behavior of the control policy in PowerGym. Specifically, we focus on the continuous battery scenario because the battery may arbitrarily discharge/charge within an allowable range in practice. For this case study, we consider four variations: vanilla (\texttt{cbat}), scaled loads (\texttt{cbat\char`_s2.5}), with soc penalty (\texttt{soc}), and scaled loads with soc penalty (\texttt{cbat\char`_soc\char`_s2.5}). As mentioned in the default environment section, both soc penalty and large load scale make the environment more challenging, as the former introduces a non-stationary reward while the latter incurs large power consumption. Thereby, we would like to see how the control policy adapts to different scenarios.

The first row of Figure~\ref{fig:123bus_err} visualizes the average switching errors of capacitors and regulators respectively. Both errors are small in most time steps across all scenarios. Hence, the policies for both capacitors and regulators only make large changes when needed while making small adjustments the rest of the time. Note the behavior of the first 1000 steps and the later steps are different because the RL exploration starts after the first 1000 steps of random exploration.

The second row of Figure~\ref{fig:123bus_err} shows the power loss ratio and the voltage violation. Because of the load scaling, the 2.5-scaled environments have the higher voltage violations than the un-scaled environments (\texttt{cbat\char`_s2.5} $>$ \texttt{cbat} and \texttt{cbat\char`_soc\char`_s2.5} $>$ \texttt{cbat\char`_soc}). Furthermore, the voltage violation of \texttt{soc\char`_s2.5} is greater than that of \texttt{s2.5} as the soc penalty makes the policy on batteries more restrictive and non-stationary. As for the power loss, since it is a difficult objective, it barely improves over time. Still, we see that the power losses on the 2.5-scaled environments are higher than the un-scaled counterparts. This is because large voltage violations cause large voltage differences on the lines, which brings up the power loss on the lines.

Finally, the third row of Figure~\ref{fig:123bus_err} shows the battery activity in discharge errors and soc errors. Since the battery is an energy storage device, it is useful when the environment lacks power and has high voltage violations. Hence, the battery barely discharges in the un-scaled environments and maintains mostly zero soc error. As for the scaled environments (\texttt{s2.5} \& \texttt{soc\char`_s2.5}), because \texttt{s2.5} discharges frequently, it has smaller voltage violations but higher soc error. In comparison, \texttt{soc\char`_s2.5} discharges less and has a higher voltage violation but smaller soc error. Therefore, there is a trade-off between battery activity and voltage violation in heavily-loaded environments: the more battery activity, the less voltage violation, and RL algorithms need to find a dedicated balance between the two.

All in all, the soc penalty and the load scale affect the difficulty of a PowerGym environment. The difficulty can be evaluated by power losses, voltage violations, and battery activities. The harder an environment is, the more power losses, voltage violations, and battery activities.

\subsection{Effects of Horizons}

\begin{figure}[!ht]
    \centering
    \includegraphics[width=0.4\textwidth]{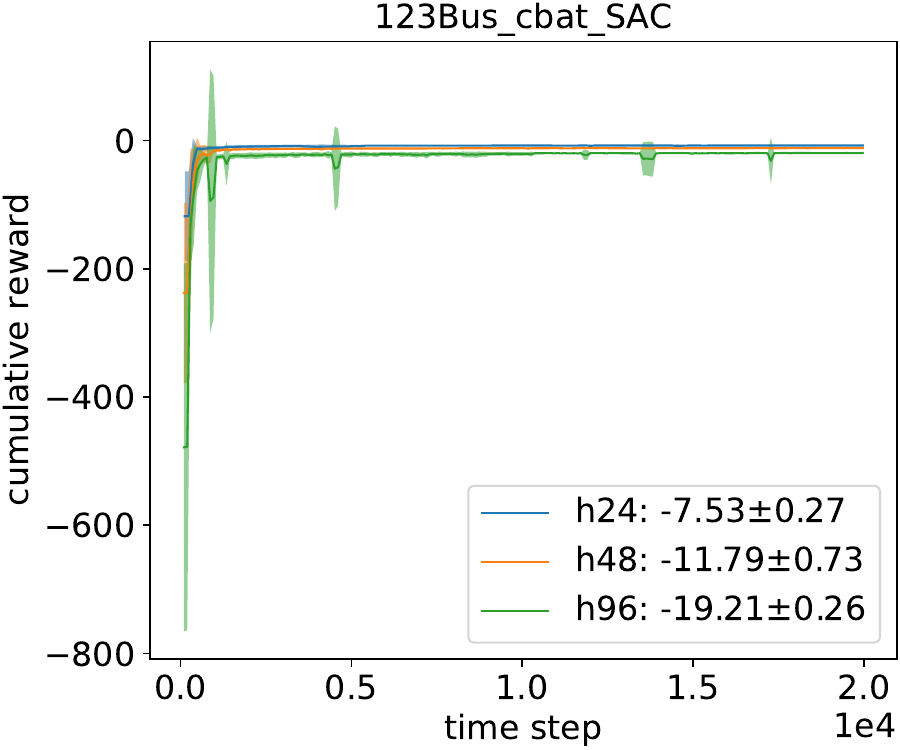}
    \includegraphics[width=0.4\textwidth]{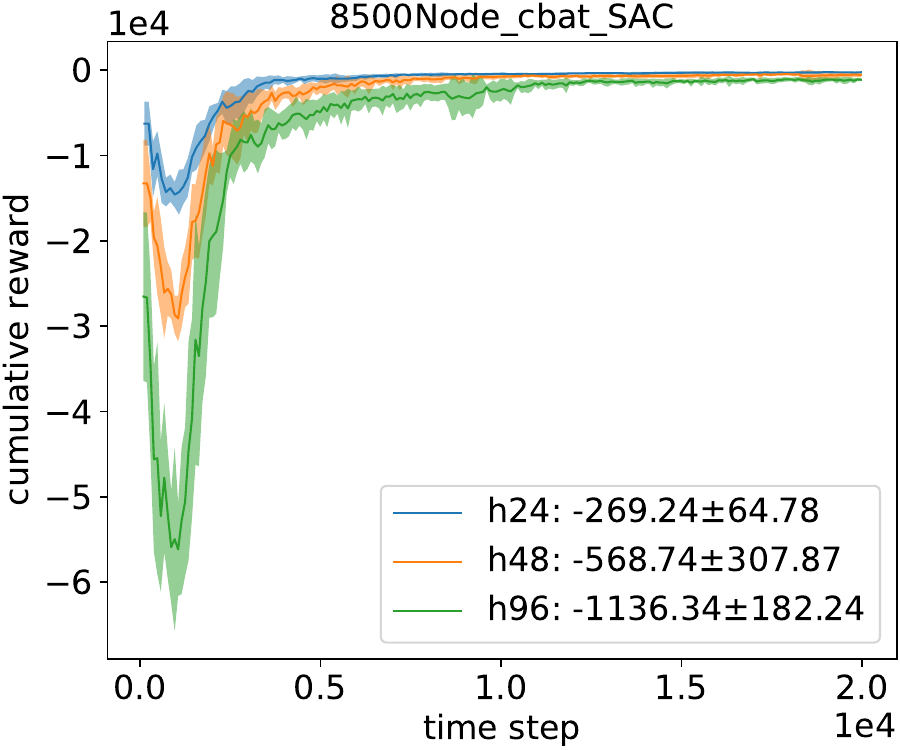}
    \caption{Effects of horizon in 20k steps. (h24, h48, h96) stand for horizon (24, 48, 96).}
    \label{fig:horizon}
\end{figure}
Figure~\ref{fig:horizon} shows the testing cumulative reward w.r.t. the horizon for 123Bus and 8500Node systems. We only analyze under the continuous battery with the SAC algorithm as this is the best-performed setting according to Figure~\ref{fig:cum_reward}. As the cumulative reward scales linearly w.r.t. the horizon, h48's cumulative reward is roughly twice of h24's and h96's is four times of h24's. Besides, the convergence speeds w.r.t. horizons are similar in 123Bus for the fact that the 123Bus system is a more stable system and less likely to have voltage violations. On the other hand, 8500Node is less stable, resulting in longer steps to converge in a longer horizon.

\subsection{Difficulty Comparisons}
As a concluding remark, we discuss the trend of PowerGym's difficulty in four aspects: problem size, base voltage violation, load scale, and soc penalty. It helps users to choose the best environment for their applications.

Problem size refers to the dimensions of an environment; e.g., horizon, sizes of state, and action spaces. The larger horizon makes an environment harder as the error of value/Q-function is usually quadratic to the horizon \citep{duan2020minimax}. Similarly, the larger state and action spaces complicate an environment. Under a fixed horizon, we expect the problem complexity in PowerGym follows 8500Node $>$ 123Bus $>$ 34Bus $>$ 13Bus.

Base voltage violation is the tendency of violating the voltage in an un-scaled environment. It depends on the structure of the distribution system and the default load profiles. One may find a small system (small number of nodes) with high base voltage violation or a big system with low base voltage violation. For instance, the tendency of base voltage violation is 8500Node $>$ 34Bus $>$ 13Bus $>$ 123Bus, as the voltage violation is the major term in the reward function and the training speed in Figure~\ref{fig:cum_reward} follows the reverse order.

Load scale affects a PowerGym environment by changing the scale of the load profiles. A high load scale brings up the load power consumption, which increases the chance of voltage violations and makes the problem harder.

The soc penalty determines the stationarity of the battery behavior. With the soc penalty, the battery behaves non-stationarily as the battery should discharge at peak hours and charge at off-peak hours. Because a non-stationary behavior is harder to train than a stationary one, the soc error of \texttt{cbat\char`_soc\char`_s2.5} in Figure~\ref{fig:123bus_err} is mostly zero but less stable.

\section{Conclusion}
We develop a gym-like open-source environment, PowerGym, to facilitate RL research/adaptation for Volt-Var control in power distribution systems. PowerGym encourages power system researchers to make fair comparisons on RL algorithms using the same environment. It includes sufficient variations (problem size, base voltage violation, load scale, and soc penalty) to study different aspects of the Volt-Var control. PowerGym also acts as a base for researchers/engineers to adopt RL algorithms to power distribution systems in real life: it provides a detailed customization guide for researchers/engineers who use PowerGym with their own proprietary power distribution systems. Our RL experiments suggest the correctness of the PowerGym design. The cumulative rewards achieved by our RL agents serve as a baseline for the PowerGym users. Future work on other problems in power distribution systems is underway.

\section{Acknowledgement}
We thank Siddharth Bhela for the instructions on OpenDSS, Suat Gumussoy for the feedbacks on environment design and Ulrich Muenz for the general supports on the project development.

\bibliographystyle{unsrtnat}
\bibliography{aaai22}

\appendix

\section{Appendix}
\subsection{System Layouts}

\begin{figure}[!ht]
    \centering
    \includegraphics[width=0.49\textwidth]{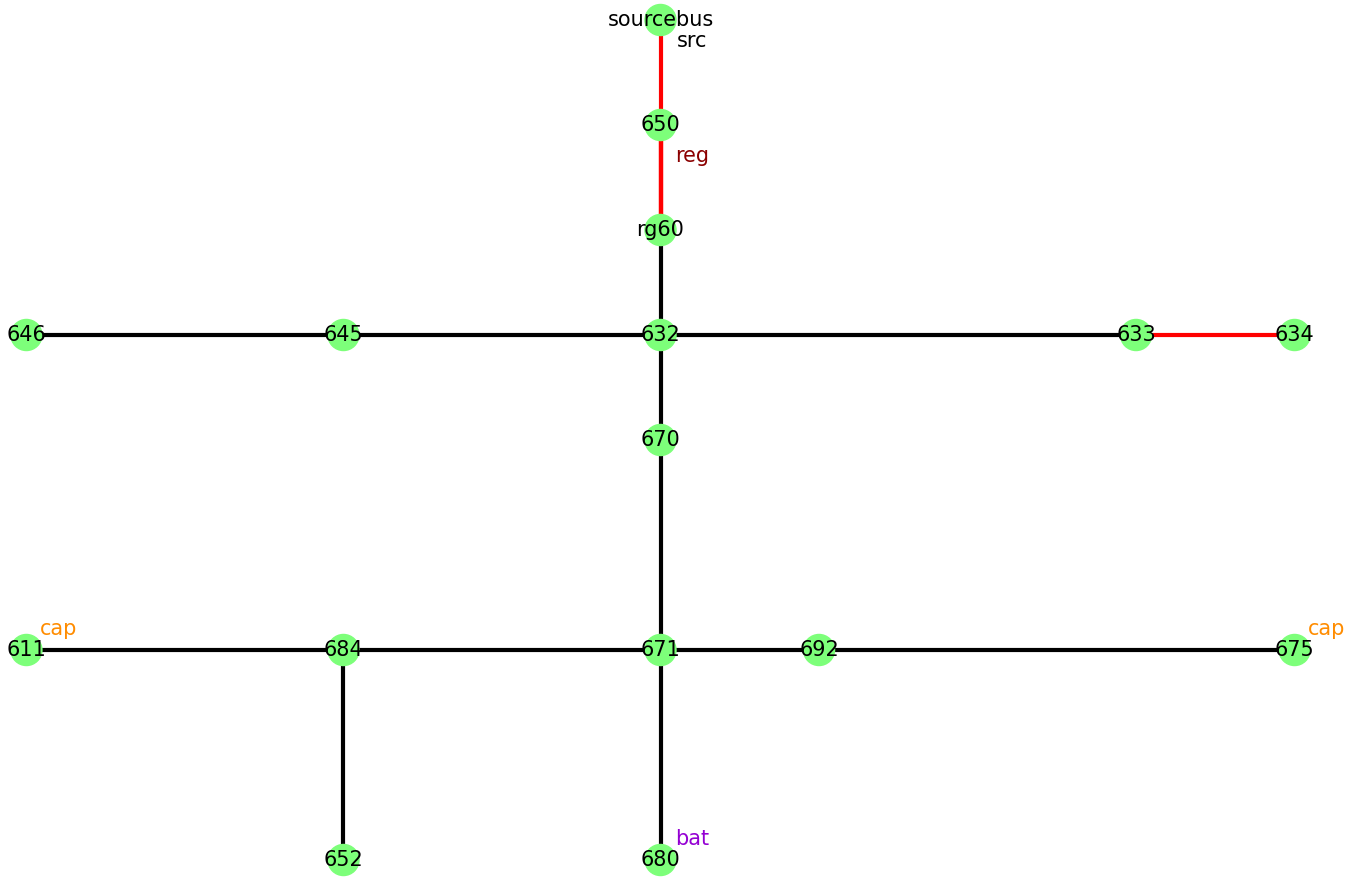}%
    \includegraphics[width=0.49\textwidth]{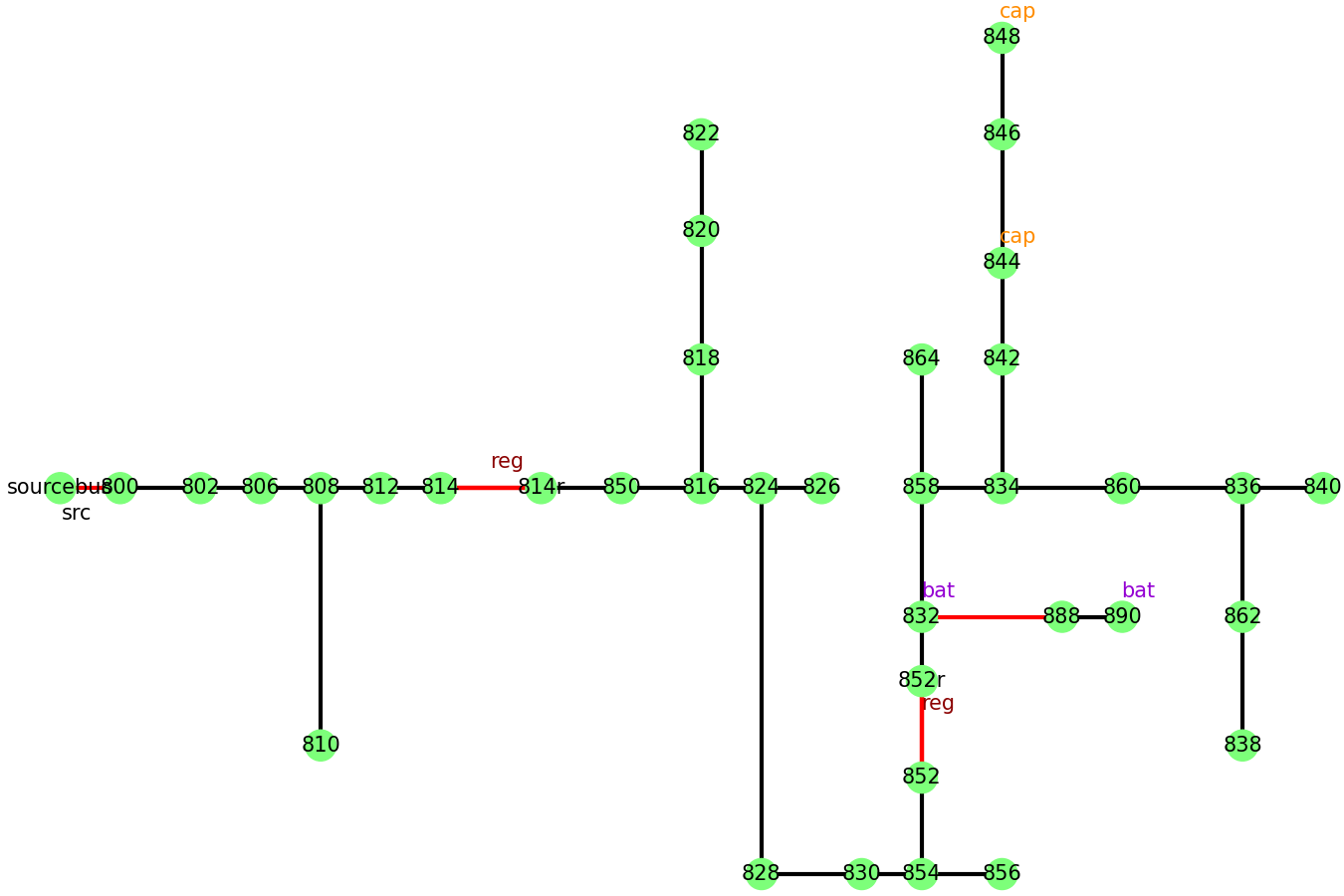}
    \includegraphics[width=0.49\textwidth]{123Bus.png}%
    \includegraphics[width=0.49\textwidth]{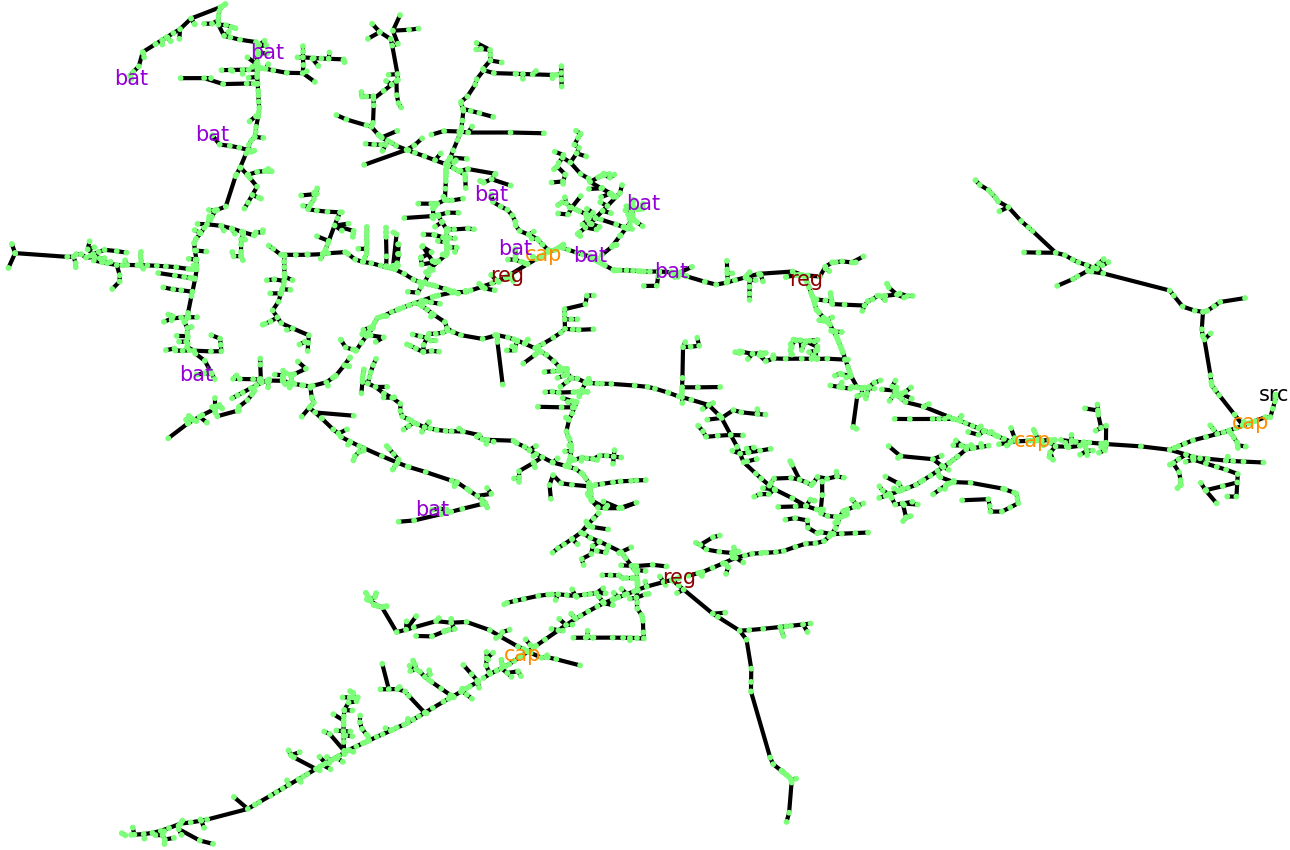}
    \caption{System layouts: 13Bus, 34Bus, 123Bus, 8500Node (top left to bottom right).}
    \label{fig:layouts}
\end{figure}

Table~\ref{tbl:sys} (in main context) and Figure~\ref{fig:layouts} show the controller summary and layouts of the systems. 13Bus system has 2 capacitors at bus 611, 675; 3 single-phase regulators at the edge (650, rg60); 1 battery at bus 680. 34Bus system has 2 capacitors at bus 844, 848; 6 single-phase regulators at edge (814,814r), (852, 852r) (3 for each); 2 batteries at bus 832, 890. 123Bus system has 4 capacitors at bus 83, 88, 90, 92; 1 three-phase regulator at bus 150 and 9 single-phase regulators at bus 9, 25, 160 (3 for each); 4 batteries at bus 33, 67, 114, 300. 8500Node system has 10 capacitors (one three-phase near the source, nine single-phase at three other locations); 12 single-phase regulators at four buses (3 for each bus. One bus is the source. The other buses are shown in the figure); 10 batteries.

\subsection{Observation and Action Wrapper}
Although the observation and action spaces are composed of discrete and continuous values, for the conciseness of representation, we wrap the observation and the action into Numpy arrays as follows.

\begin{quote}
\begin{verbatim}
wrapped_obs = Concatenate([all phase voltages at each bus, 
                           all capacitor statuses, 
                           all regulator tap numbers, 
                           all battery soc's and normalized discharge powers])
\end{verbatim}    
\end{quote}

\begin{quote}
\begin{verbatim}
action = Concatenate([all capacitor statuses, 
                      all regulator tap numbers, 
                      all battery discharge powers])
\end{verbatim}
\end{quote}
Capacitor statuses, regulator tap numbers, and discrete batteries' discharge powers are represented in integers. Continuous batteries' discharge powers are represented in floating numbers.

The wrapped observation is the default output of \texttt{Env.reset()} and \texttt{Env.step()}. Still, users can access all phase voltages with the observation dictionary at \texttt{Env.obs}. These two representations of observations have the following relation:
\begin{quote}
\begin{verbatim}
wrapped_obs = Env.wrap_obs(Env.obs)    
\end{verbatim}
\end{quote}

\subsection{Hyper-parameters}

In this section, we provide a summary of the hyper-parameters of our environments and RL agents. The coefficients of the reward function are shown in Table~\ref{tbl:reward_coeff}.

\begin{table}[!ht]
    \centering
    \begin{tabular}{l|cccc}
        \hline
        \hline
        \textbf{Variable} & \textbf{13Bus} & \textbf{34Bus} & \textbf{123Bus} & \textbf{8500Node} \\ \hline
        $N_{\text{reg\_act}}$ & \multicolumn{4}{c}{33} \\
        $N_{\text{bat\_act}}$ & \multicolumn{4}{c}{33 (disc. bat), $\infty$ (cont. bat)} \\
        $\text{w}_{\text{cap}}$ & \multicolumn{4}{c}{1/33} \\
        $\text{w}_{\text{reg}}$ & \multicolumn{4}{c}{1/33} \\
        $\text{w}_{\text{power}}$ & 10.0 & 1.0 & 10.0 & 1.0 \\
        $\text{w}_{\text{dis}}$ & 6/33 & 10/33 & 7/33 & 200/33\\
        $\text{w}_{\text{soc}}$ & \multicolumn{4}{c}{0.0}\\
        \hline
        \hline
    \end{tabular}\\
    ~\newline
    ~\newline
    \begin{tabular}{l|cccc}
        \hline
        \hline
        \textbf{Variable} & \textbf{13Bus} & \textbf{34Bus} & \textbf{123Bus} & \textbf{8500Node} \\ \hline
        $\text{w}_{\text{dis}}$ & 1/33 & 4/33 & 5/33 & 200/33\\
        $\text{w}_{\text{soc}}$ & 100/33 & 500/33 & 500/33 & 10000/33\\
        \hline
        \hline
    \end{tabular}
    \caption{Reward Coefficients: without soc (top), with soc (bottom). The two settings only differ in $\text{w}_{\text{dis}}$ and $\text{w}_{\text{soc}}$, so the RHS only presents their values.}
    \label{tbl:reward_coeff}
\end{table}

To train PPO and SAC agents, we use separate deep neural networks to parameterize the policy and value/Q functions. Both networks consist of dense layers with the same widths. Table~\ref{tbl:params} presents the suggested hyper-parameters for PPO and SAC.

\begin{table}[!ht]
    \centering
    \begin{tabular}{l|c}
        \hline
        \hline
        \textbf{Variable} & \textbf{Value} \\ \hline
        Optimizer & Adam \\
        Learning rate & 3E-4 \\
        Discount factor & 0.95 \\
        Clip epsilon & 0.2 \\
        Batch size & 64 \\
        Model update interval & 512 \\
        Entropy coefficient & 0.01 \\
        \hline
        \hline
    \end{tabular}
    \quad
    \begin{tabular}{l|c}
        \hline
        \hline
        \textbf{Variable} & \textbf{Value} \\ \hline
        Optimizer & Adam \\
        Learning rate & 3E-4 \\
        Discount factor & 0.95 \\
        Batch size & 256 \\
        Model width & 512 \\
        Model depth & 3 \\
        Entropy coefficient & 0.4 \\
        \hline
        \hline
    \end{tabular}
    \caption{Suggested hyper-parameters for PPO (left) and SAC (right).}
    \label{tbl:params}
\end{table}

\end{document}